\documentclass{article}

\usepackage[T1]{fontenc}
\ifdefined\XeTeXversion
  \catcode"2013=\active
  \def–{\textendash}
\fi
\usepackage{arxiv}
\usepackage[authoryear,round]{natbib}
\renewcommand{\headeright}{}

\renewcommand{\shorttitle}{}
\date{}
\usepackage{amsmath,amssymb,mathtools}
\usepackage{graphicx}
\usepackage{booktabs}
\usepackage{tabularx}
\usepackage{multirow}
\usepackage{array}
\usepackage{enumitem}
\usepackage{microtype}
\usepackage{xcolor}
\usepackage[hypertexnames=false]{hyperref}
\usepackage[capitalize,noabbrev]{cleveref}
\usepackage{placeins}
\usepackage{siunitx}

\usepackage{xspace}

\hypersetup{colorlinks=true,citecolor=blue!65!black,linkcolor=blue!65!black,urlcolor=blue!65!black}
\newcommand{\method}{\textsc{DeXplicit}\xspace}

\newcommand{\R}{\mathbb{R}}

\title{DeXplicit: Making Physical State Explicit in Dexterous Manipulation World Models
}
\author{Adam Lee$^{1,2}$\thanks{Work done during an internship at S-Lab.} \quad Mutian Xu$^{1,\dagger}$ \quad Yukang Cao$^{1}$ \quad Shulin Tian$^{1}$ \quad Ziwei Liu$^{1,\ddagger}$\\
{\normalfont\small $^{1}$S-Lab, Nanyang Technological University \quad $^{2}$University of California, Berkeley}\\
{\normalfont\small $^{\dagger}$Project Lead \quad $^{\ddagger}$Corresponding Author}}

\begin{document}
\maketitle
\lhead{\footnotesize DeXplicit}

\begin{abstract}
    Before acting on an object, a robot needs to anticipate how its hand motion will change the physical world and what the resulting scene will look like. Existing approaches typically entangle action-conditioned dynamics, camera motion, and visual synthesis within a video generation model, or represent dynamics as point trajectories without explicit object structure. We introduce \method, a dexterous manipulation world model that predicts how the world evolves explicitly in metric 3D before rendering its visual consequences. Given an initial scene and a future hand trajectory, a lightweight 3D dynamics network predicts object-centric motion and deformation, providing structured and geometrically grounded future world states. These predictions guide video generation through appearance transport and geometric projection, allowing the renderer to focus on uncertain and newly visible content rather than implicitly learning object dynamics from pixels.
    To retain these controls in an efficient autoregressive generator, we develop a two-stage causal adaptation procedure: alignment of the transferred geometry, hand, and appearance conditions to a frozen few-step backbone, followed by on-policy distillation with a conditional video critic. The resulting model generates four-frame latent blocks with two denoising steps while conditioning on its generated history.
    Across HOT3D, OakInk2, and PhysTwin, \method reduces moved-point error by 27--40\% relative to PointWorld and improves object IoU$_+$ by 47\% on HOT3D under a shared renderer. Causal-\method retains an object IoU$_+$ of 0.490 while accelerating video synthesis by 57 times relative to our 14B bidirectional configuration.
    Project page: \href{https://dexplicit.github.io}{https://dexplicit.github.io}.
    
\end{abstract}

\begin{figure}[htbp]
\centering
\includegraphics[width=0.8\linewidth]{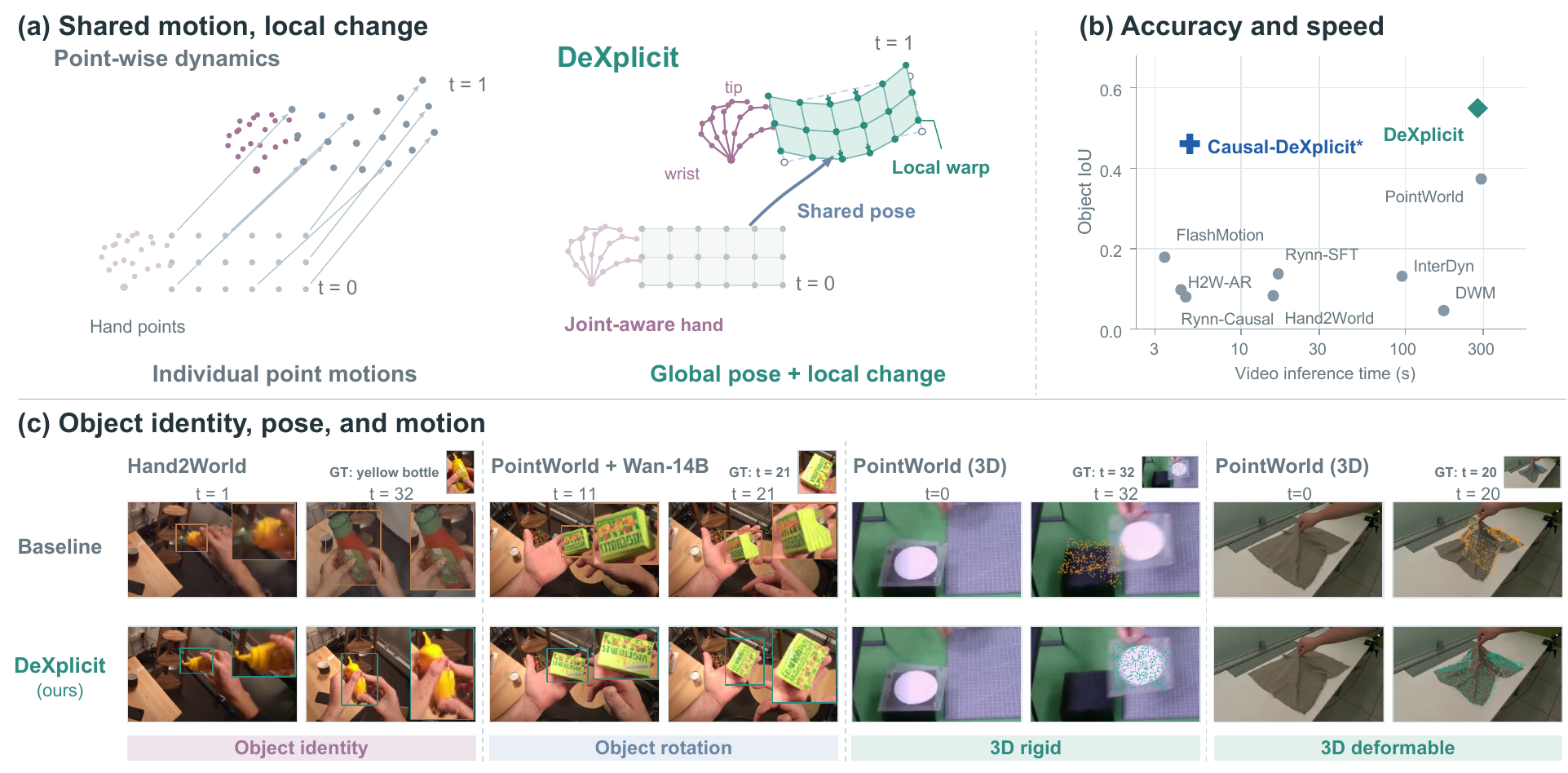}
\caption{\textbf{Explicit physical state for dexterous interaction.} Structured interaction dynamics connect metric prediction to video synthesis, preserving transport, articulation, and deformation. The latency plot compares native video systems, with the asterisk denoting the Fast Adapter starting model. Protocols are detailed in \Cref{sec:hot48}.}
\label{fig:teaser}
\end{figure}
\section{Introduction}
World models for dexterous manipulation increasingly use action-conditioned video generation~\citep{ha2018worldmodels,hafner2023dreamerv3,hansen2024tdmpc2,yang2024unisim,zhou2024robodreamer,zhu2024irasim,nvidia2025cosmos} to predict the visual consequences of an agent's actions.
However, visual plausibility does not ensure that an object responds correctly to the action or preserves its identity throughout the interaction. 
This gap is especially important in dexterous manipulation, where small errors in object pose or local geometry can break hand-object contact relationships.

A key difficulty in using action-conditioned video generation for this task is that pixels entangle the object's physical evolution only with its visual appearance~\citep{du2023unipi,wu2023gr1,zhou2026ctrlworld,kim2025dwm,goswami2025dexwm,wang2026hand2world}.
A model may synthesize a convincing grasp while placing the object at an incorrect 3D pose, or allow its geometry and surface features to drift through occlusion.
These failures reflect the burden of leaving physical state implicit: the generator must learn how hand actions change the object while simultaneously determining how those changes appear under varying viewpoints and occlusion~\citep{sudhakar2024coshand,akkerman2024interdyn,wang2026hand2world,kim2025dwm,goswami2025dexwm}.

Metric 3D prediction~\citep{huang2026pointworld,huang2025particleformer,zhang2025pgnd,duisterhof2026pointzero,soraki2026objectforesight} provides this missing interface by representing the object's physical evolution as an explicit intermediate state for visual synthesis.
PointWorld expresses actions and scene evolution as point trajectories in a shared coordinate system~\citep{huang2026pointworld}, making motion measurable before rendering.
However, this generic point trajectory does not explicitly encode the distinct structures of the hand and the manipulated object~\citep{goswami2025dexwm,yuan2026dexac,wang2026hand2world,xie2026generatedreality,kim2025dwm}.
In physical hand–object interactions, hand joints are linked by a kinematic hierarchy, whereas object points share coherent motion with local changes caused by articulation or deformation.
Leaving these relationships implicit~\citep{sanchezgonzalez2020gns,pfaff2021meshgraphnets,huang2025particleformer} can allow object geometry to distort or drift relative to the interacting hand, compromising both physical prediction and subsequent visual synthesis.

To address these limitations, we introduce \method (\Cref{fig:teaser}), a dexterous manipulation world model that explicitly predicts how the 3D world evolves before generating visual consequences. Our key idea is to \textit{separate physical state prediction from visual synthesis}, rather than requiring a single heavy video model to implicitly learn action-conditioned 3D dynamics, camera motion, and appearance generation altogether.
\textbf{1) Explicit 3D state prediction.}
Given an initial scene and a future hand trajectory, we predict the resulting geometric changes directly in metric 3D space using a lightweight dynamics network. By reasoning over object-centric motion and deformation, the model explicitly captures how hand actions transform the world state, providing geometrically grounded and directly interpretable future states.
\textbf{2) 3D-guided video readout.}
Once the future 3D states are predicted, video generation becomes a substantially easier rendering problem rather than implicit world-dynamics prediction. We propagate appearance through the predicted 3D motion and use it as a strong visual and geometric scaffold for generation. This explicitly accounts for object motion and naturally incorporates viewpoint changes through 3D projection, allowing the video model to focus primarily on completing uncertain or newly visible content. Building on this decomposition, we develop a geometry-conditioned causal video generator through a two-stage adaptation procedure. We first align the learned geometry, hand, and appearance-transport conditions with a pretrained few-step causal backbone while freezing its generative parameters. We then apply on-policy autoregressive distillation with a conditional video critic, allowing the model to learn from its own generated history without changing the two-step sampling procedure. This preserves the explicit physical-state interface while substantially reducing visual synthesis cost.

We evaluate \method on HOT3D, OakInk2, and PhysTwin~\citep{banerjee2025hot3d,yang2024oakink2,jiang2025phystwin}, covering rigid, articulated, and deformable interactions.
Compared with PointWorld, \method reduces long-horizon moved-point error by $39.9\%$, $28.0\%$, and $27.1\%$ on the three datasets, respectively.
These improvements also translate into more accurate object response in generated videos: under the same renderer, replacing PointWorld predictions with ours increases object IoU$_+$ from 0.373 to 0.549 on HOT3D.
Our geometry-conditioned causal adaptation further enables 32-frame synthesis in 4.90\,s on a single NVIDIA A800 GPU, while retaining an object IoU$_+$ of 0.490 under two-step autoregressive generation.
Together, these results establish structured physical-state prediction as a key driver of interaction fidelity in dexterous world models, linking accurate action-induced object motion to faithful visual outcomes.

\section{Related Work}
\paragraph{World models for interaction.}
Learned latent dynamics support imagination and control~\citep{ha2018worldmodels,hafner2023dreamerv3,hansen2024tdmpc2}, while UniPi and GR-1 connect video prediction to robot policies~\citep{du2023unipi,wu2023gr1}. UniSim and RoboDreamer broaden visual simulation through interactive and compositional prediction~\citep{yang2024unisim,zhou2024robodreamer}. IRASim and Ctrl-World emphasize action-conditioned robot interaction~\citep{zhu2024irasim,zhou2026ctrlworld}, supported by scalable video priors such as Cosmos~\citep{nvidia2025cosmos}. Together, these models connect visual prediction to embodied control.

\paragraph{Dexterous and egocentric world models.}
CosHand, InterDyn, and Mask2IV use silhouettes or masks to specify interaction~\citep{sudhakar2024coshand,akkerman2024interdyn,li2026mask2iv}. DWM combines articulated hand control with scene reconstruction~\citep{kim2025dwm}. Hand2World and Generated Reality model egocentric interaction from hand and camera inputs~\citep{wang2026hand2world,xie2026generatedreality}, while HandsOnWorld disentangles these controls and RynnWorld enables digital teleoperation~\citep{chen2026handsonworld,zhao2026rynnworld}. EgoSim maintains an updatable 3D scene state~\citep{hao2026egosim}. DexWM and DexAC-WM emphasize hand-consistent latent prediction and structured action conditioning~\citep{goswami2025dexwm,yuan2026dexac}. We focus on measurable object dynamics, coupling metric forecasts of transport and deformation to visual synthesis.


\paragraph{Explicit physical dynamics.}
Particle and mesh simulators learn spatial interactions through graph networks~\citep{sanchezgonzalez2020gns,pfaff2021meshgraphnets}. ParticleFormer and particle-grid neural dynamics extend learned simulation to perceived objects and materials~\citep{huang2025particleformer,zhang2025pgnd}. PointWorld unifies scene and action trajectories in point space~\citep{huang2026pointworld}, and PointZero learns transferable dynamics through track completion~\citep{duisterhof2026pointzero}. ObjectForesight predicts object poses from human video, while RigidFormer imposes coherent rigid motion~\citep{soraki2026objectforesight,dou2026rigidformer}. Our shared object pose captures transport, and an embedded graph~\citep{sumner2007embedded} expresses residual deformation, preserving coherence without restricting all objects to rigid motion.

\paragraph{Controllable video generation.}
ControlNet and VACE introduce spatial conditioning~\citep{zhang2023controlnet,jiang2025vace}, while MotionCtrl, CameraCtrl, and GEN3C address motion and viewpoint control~\citep{wang2023motionctrl,he2024cameractrl,ren2025gen3c}.
ManiVideo and PAM incorporate hand--object geometry and appearance, and ContactFlow transfers contact-based controls across embodiments~\citep{pang2024manivideo,gao2026pam,azirar2026contactflow}. DVG-WM separates interaction prediction from visual refinement~\citep{shan2026dvgwm}. Closest to our appearance-transport interface, WanMove transports initial features and E3C maintains 3D visual memory~\citep{chu2025wanmove,gu2026e3c}. We couple initial appearance to persistent object points, using depth, visibility, and forecast confidence to determine valid support.



\section{Problem Formulation}
\label{sec:formulation}

We study action-conditioned dexterous manipulation: given an initial scene and proposed hand trajectory, the model predicts the resulting metric 3D state and corresponding future observations while preserving the object's response, identity, and appearance.


\paragraph{Inputs.}
The inputs are the initial image $I^0$, metric scene points $X^0 = \{x_i^0\}_{i=1}^{N}$, anatomically indexed hand trajectory $H^{0:T}$, and calibrated cameras $\Pi$. 
Each object point $x_i$ has an object label $o_i$ and canonical coordinate $c_i$ in the object's reference shape.


\paragraph{Physical prediction and visual synthesis.}
We decompose the world model as a dynamics predictor $F_\theta$, a deterministic projection operator $\mathcal R$, and a conditional video renderer $G_\psi$:
\begin{equation}
\begin{aligned}
 (\widehat X^{1:T},Q^{1:T})
     &=F_\theta(X^0,H^{0:T}),\\
 (C^{0:L},M^{0:L})
     &=\mathcal R(I^0,\widehat X,Q,H,\Pi),\\
 \widehat I^{1:L}
     &=G_\psi(I^0,C,M,\epsilon),
\end{aligned}
 \label{eq:factor}
\end{equation}
where $\widehat X^{1:T}$ contains predicted metric point positions and $Q^{1:T}=\{q_i^t\}$ their confidences, with $q_i^t\in[0,1]$. The projection operator $\mathcal R$ constructs visual controls $C$ and soft mask $M$ from geometry, confidence, appearance, hand poses, and cameras to condition $G_\psi$. 

\section{\method}
\label{sec:method-dexsim}

\begin{figure}[t]
\centering
\includegraphics[width=\linewidth]{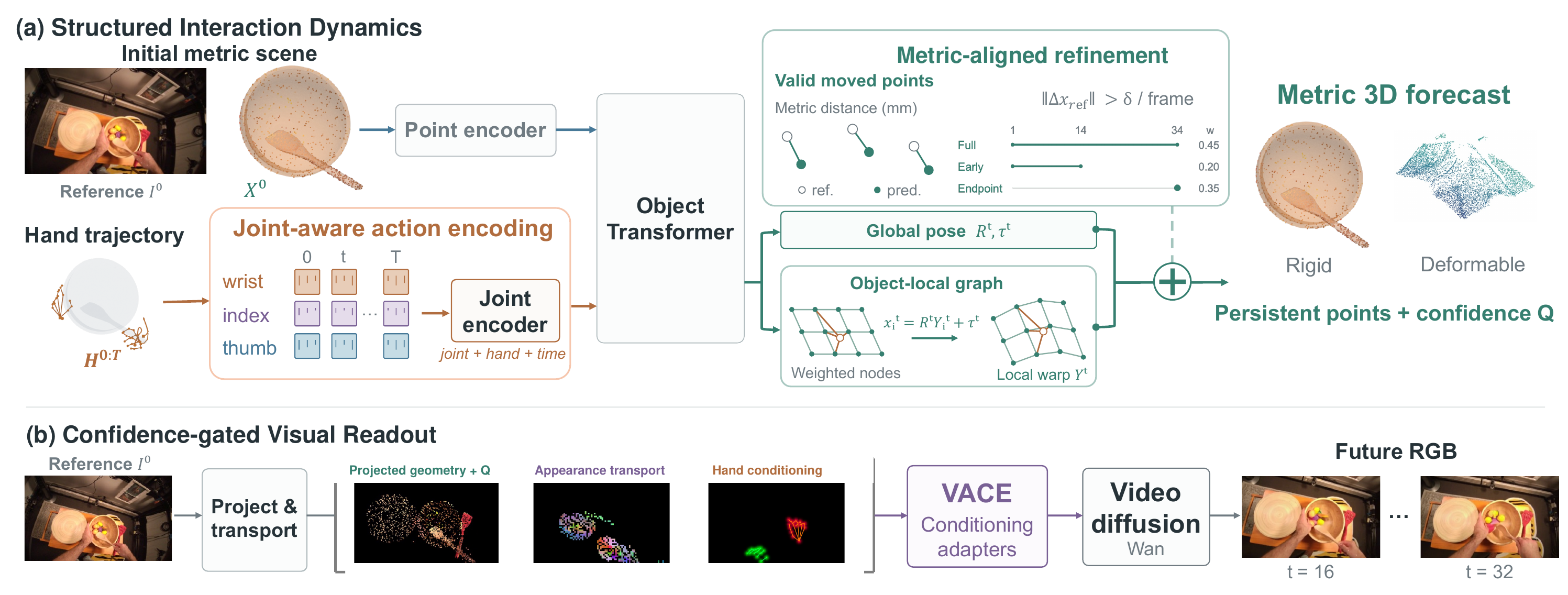}
\caption{\textbf{Overview of DeXplicit.} Joint-aware action encoding drives object-centric metric dynamics with shared object transport and local deformation. Metric-aligned refinement corrects object pose, and projected geometry, hand conditioning, and appearance transport condition the confidence-gated visual readout.}
\label{fig:method-architecture}
\end{figure}


\subsection{Structured Interaction Dynamics}
\label{sec:structured}

Representing hand actions and object motion as generic point trajectories leaves joint relationships and object-level coherence implicit.
We therefore structure the dynamics predictor $F_{\theta}$ in Sec.~\ref{sec:formulation} around joint-aware action encoding and object-centric motion prediction by encoding the hand's anatomical structure and predicting object motion through shared global transport and local deformation (\Cref{fig:method-architecture}).

\paragraph{Joint-aware action encoding.}
For joint $j$ of hand $h\in\{L,R\}$ at time $t$, let $p_{t,h,j}\in\R^3$ denote its position, with joint 0 representing the wrist and $\pi(j)$ its kinematic parent. We construct:
\begin{equation}
    \begin{aligned}
        a_{t,h,j}
        &=
        \phi_{\mathrm{joint}}\!\left(
        \left[
        p_{t,h,j},\;
        p_{t,h,j}-p_{t,h,0},\;
        \Delta p_{t,h,j},\;
        b_{t,h,j},\;
        \|b_{t,h,j}\|_2,\;
        v_{t,h,j}
        \right]
        \right)
        +e_j+e_h+e_t,\\
        A
        &=
        E_{\mathrm{hand}}\!\left(
        \{a_{t,h,j}\},\;
        \{m_{t,h}\},\;
        \Delta H^{0:T}
        \right).
    \end{aligned}
\label{eq:joint-aware-action}
\end{equation}
Here, $p_{t,h,j}-p_{t,h,0}$ denotes the wrist-relative position, $\Delta p_{t,h,j}$ is the displacement from the preceding time step, and $b_{t,h,j} = p_{t,h,j}-p_{t,h,\pi(j)}$ describes the bone connecting the joint to its parent.
The validity flag $v_{t,h,j}$ identifies available observations.
A learned projection $\phi_{\mathrm{joint}}$ embeds these geometric features, while $e_j$, $e_h$, and $e_t$ encode joint identity, hand side, and time.
The hand-level features $m_{t,h}$ contain MANO pose, wrist transform, and shape when available. 
Finally, $\Delta H^{0:T}=\{\Delta H_t\}_{t=0}^{T}$ denotes the sequence of action-state differences, where $\Delta H_t=H_t-H_{t-1}$ and $\Delta H_0=0$.

\paragraph{Object-centric metric dynamics.}
To predict its effect on the scene, we encode the initial points with a Point Transformer and let their features attend to $A$.
To encourage coherent motion within each object, we aggregate these features by object using mean and max pooling and combine them with canonical coordinate information to form object tokens $Z$.
An object-time Transformer combines $Z$ with the encoded hand action to produce an object-time feature $U_o^t$ for each object $o$ and future time $t$.
A pose output network maps $U_o^t$ to a shared global rotation $R_o^t$ and translation $\tau_o^t$.

To also represent articulation and deformation, we place $K$ control nodes $g_{o,k}$ on each object's canonical geometry.
For each node, we combine the object-time feature $U_o^t$ with features pooled from nearby points, the node's canonical position, and its spatial relationships to the hand joints.
A deformation network maps these features to a local rotation $r_{o,k}^t$ and translation $d_{o,k}^t$.
These local transforms deform the canonical points before the global
object pose is applied:
\begin{equation}
\begin{aligned}
    Y_i^t
    &=
    \sum_{k=1}^{K}w_{ik}
    \left[
        r_{o_i,k}^t(c_i-g_{o_i,k})
        +g_{o_i,k}+d_{o_i,k}^t
    \right],\\
    \widehat{x}_i^t
    &=R_{o_i}^tY_i^t+\tau_{o_i}^t.
\end{aligned}
\label{eq:object-centric-dynamics}
\end{equation}
The fixed normalized weights $w_{ik}$ determine how strongly nearby nodes influence point $i$.

\paragraph{Bounded pose correction.}
The initial object pose can remain inaccurate even when its points move coherently.
We therefore add a lightweight adapter that predicts a small adjustment to each pose.
The adapter outputs a twist $(v,\omega)$ in the object's local frame, with componentwise bounds of 5\,cm and $15^\circ$.
The $\mathrm{SE}(3)$ exponential map converts it into a rigid correction:
\begin{equation}
    R_\Delta=\exp([\omega]_\times),
    \qquad
    t_\Delta=J(\omega)v,
    \label{eq:pose-correction}
\end{equation}
where $[\omega]_\times$ is the skew-symmetric matrix corresponding to $\omega$, and $J(\omega)$ is the left Jacobian of $\mathrm{SO}(3)$.
We compose the correction with the initial pose:
\begin{equation}
    R :=RR_\Delta,
    \qquad
    \tau :=\tau+Rt_\Delta.
    \label{eq:pose-composition}
\end{equation}
The adapter's output layer is zero-initialized, so adding it initially
leaves the prediction unchanged.

Pointwise uncertainty is estimated from the predicted trajectories and converted into confidence $Q^{1:T}$ for visual synthesis. The following section describes the geometric supervision and pose refinement used to improve the position predictions.

\begin{figure}[!t]
\centering
\includegraphics[width=\linewidth]{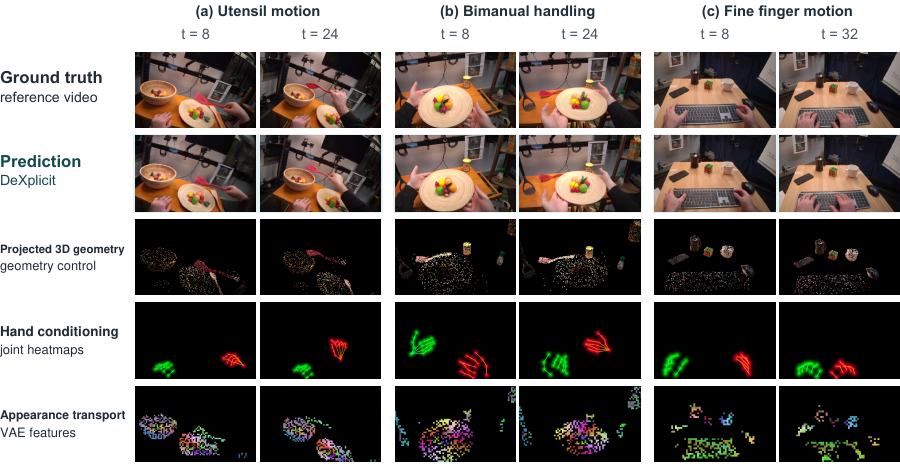}
\caption{\textbf{Conditioning interface for video generation.} Projected 3D geometry, hand conditioning, and appearance transport provide complementary controls. Examples show the conditions and generated results at matched times and crops. The appearance-transport row visualizes transported initial-image VAE features; colors use a fixed PCA basis, black indicates unavailable appearance, and skeletons visualize joint heatmaps.}
\label{fig:wan13b-conditions}
\end{figure}

\subsection{Metric-aligned Refinement}
\label{sec:refinement}
We therefore introduce metric-aligned refinement to directly reduce object-point position errors and use hand-derived motion references to guide bounded pose corrections.

\paragraph{Metric-aligned supervision.}
To emphasize action-induced object motion, we introduce a moved-point objective that measures geometric errors on moving regions:
\begin{equation}
    E_{\mathcal{T}}^{\mathrm{mov}}
    =
    \frac{
        \sum_{w,i,\,t\in\mathcal{T}}
        m_{w,i}^{t}v_{w,i}^{t}
        \|\widehat{x}_{w,i}^{t}-x_{w,i}^{t}\|_2
    }{
        \sum_{w,i,\,t\in\mathcal{T}}
        m_{w,i}^{t}v_{w,i}^{t}
    }.
    \label{eq:metric}
\end{equation}
Here, $w$ indexes sequence windows, $\mathcal{T}$ specifies the future states included in the loss, and $v_{w,i}^t$ marks valid targets.
The motion mask $m_{w,i}^t$ selects points whose reference displacement from the preceding state exceeds 5\,mm.
We also define $E_{34}^{\mathrm{ever}}$ as the endpoint error over points that move at any time in the sequence.

Using smoothed training versions $\widetilde{E}$ of these errors, we combine full-horizon, early-horizon, and endpoint supervision:
\begin{equation}
    \mathcal{L}_{\mathrm{metric}}
    =
    0.45\,\widetilde{E}_{1:34}^{\mathrm{mov}}
    +
    0.20\,\widetilde{E}_{1:14}^{\mathrm{mov}}
    +
    0.35\,\widetilde{E}_{34}^{\mathrm{ever}}.
    \label{eq:scoreloss}
\end{equation}

\paragraph{Hand-derived motion references.} 
We guide bounded pose correction with candidate object poses derived from rigid motion of the left hand, right hand, or both hands, together with static and current-prediction hypotheses; the model learns from their differences to refine the object pose.

\subsection{Confidence-gated Visual Readout}
\label{sec:visual}
To provide the renderer with the corresponding appearance, we attach features from the initial image to object points and project the confidence values into a soft edit mask, guiding the renderer to retain reliable content and synthesize uncertain or uncovered regions (\Cref{fig:wan13b-conditions}).

\paragraph{Appearance transport.}
Following appearance-transport approaches~\citep{chu2025wanmove,gu2026e3c}, we encode the initial image $I^0$ with a frozen VAE to obtain a feature map $f^0$.
For $x_i^0$, we sample an appearance descriptor at its projected location $u_i^0=\Pi^0(x_i^0)$:
\begin{equation}
    f_i=\operatorname{bilinear}(f^0,u_i^0).
\end{equation}
We therefore assign each object point a source-appearance indicator $a_i^0\in\{0,1\}$.
It equals 1 if the point projects inside the initial image, is not occluded and is closest to the camera among points projecting to the same feature-map cell; otherwise, it equals to 0.

At each time step $t$, we project the predicted positions $\widehat x_i^t$ into the target view while retaining their initial descriptor $f_i$ and source-appearance indicator $a_i^0$.
Let $n^*(t,u)$ denote the nearest projected point with positive depth at target location $u$.
Its source-appearance indicator determines whether an initial-image feature is available there:
\begin{equation}
    B^t(u)=
    \begin{cases}
        a_{n^*(t,u)}^0, & n^*(t,u)\text{ exists},\\
        0,  &\text{otherwise},
    \end{cases}\\
    F^t(u)=
    \begin{cases}
        f_{n^*(t,u)}, & B^t(u)=1,\\
        0, & B^t(u)=0.
    \end{cases}
\label{eq:appearance-transport}
\end{equation}
Here, $B^t$ records appearance coverage and $F^t$ contains the transported appearance features.
A causal encoder then maps $F^t$ and $B^t$ into video conditions.

\paragraph{Confidence-gated synthesis.}
Alongside appearance transport, we splat the predicted point confidence through the same camera projection to obtain $q_{\mathrm{proj}}^t(u)$, assigning zero confidence where no point projects.
This defines the soft edit mask
\begin{equation}
    M^t(u)=1-q_{\mathrm{proj}}^t(u).
    \label{eq:confidence-mask}
\end{equation}
Low mask values encourage retention of reliable projected content; high values permit synthesis in geometrically uncertain regions.
Together with projected scene and hand geometry, these signals guide the renderer to preserve reliable appearance and complete uncertain or unseen regions.

\paragraph{Conditioning-interface training.}
We train the conditioning branches and VACE adapters~\citep{jiang2025vace} using flow matching~\citep{lipman2023flowmatching,liu2023rectifiedflow}, keeping the pretrained video backbone and VAE frozen. A 14B bidirectional renderer provides high-quality synthesis, while a separate 1.3B model learns the conditioning interface used for causal adaptation.

\subsection{Geometry-Conditioned Causal Generation}
\label{sec:fast-adaptation}
\begin{figure}[t]
\centering
\includegraphics[width=\linewidth]{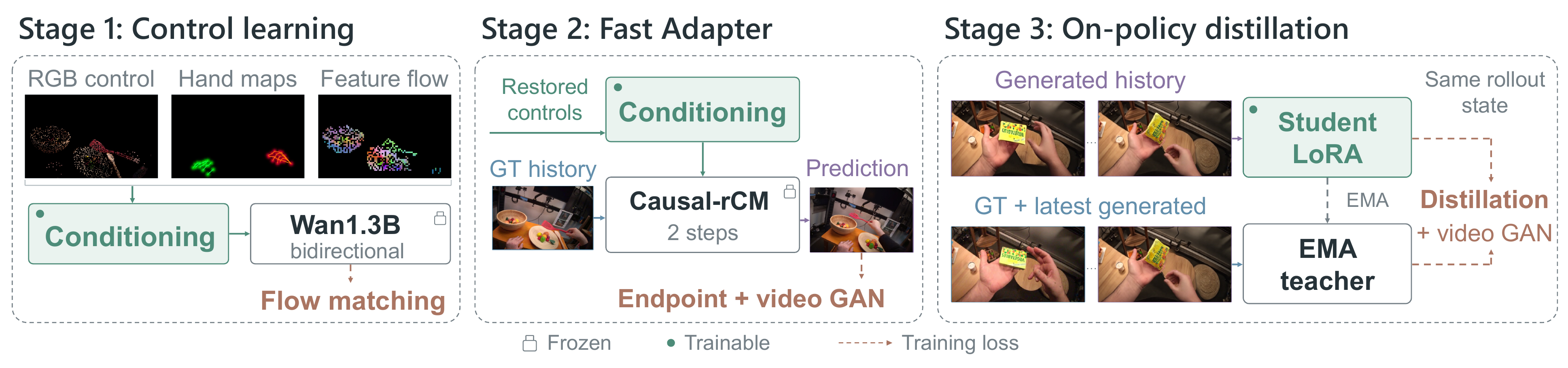}
\caption{\textbf{Sequential learning of interaction controls and causal rollout.} The conditioning interface is learned with a bidirectional renderer, aligned to a frozen few-step causal model, and then refined through on-policy adaptation.}
\label{fig:multi-stage}
\end{figure}

The conditioning interface in Sec.~\ref{sec:visual} specifies object geometry, hand motion, and transported appearance independently of the video backbone. To retain these controls in a few-step autoregressive renderer, we develop a two-stage causal adaptation procedure (\Cref{fig:multi-stage}). We first transfer and align the conditioning parameters learned with the separate 1.3B bidirectional model to a frozen Causal-rCM backbone~\citep{zheng2026causalrcm}. We then adapt the conditioned generator on its own autoregressive rollouts using on-policy distillation and a conditional video critic. These stages acquire interaction control and improve generated-history consistency while preserving the backbone's two-step sampling procedure.

\paragraph{Condition transfer and alignment.}
Let $z_{0,s}$ denote the target latent for block $s$, and let $C_{\leq s}$ contain the geometry, hand, and appearance conditions available through that block. With frozen backbone parameters $\theta_0$ and trainable conditioning parameters $\phi$, we predict a clean endpoint from a noisy target latent at a deployed noise level $\sigma$:
\begin{equation}
\begin{aligned}
 \widehat z_{0,s}
 &=z_{\sigma,s}-\sigma v_{\theta_0,\phi}
 \big(z_{\sigma,s},\sigma;\operatorname{sg}(z_{0,<s}),I^0,C_{\leq s}\big),\\
 \mathcal L_{x_0}&=\frac{1}{S}\sum_{s=1}^{S}\|\widehat z_{0,s}-z_{0,s}\|_2^2.
\end{aligned}
\label{eq:fast-endpoint}
\end{equation}
Here $\operatorname{sg}$ stops gradients through the clean historical latents and $S=8$ future blocks. Endpoint reconstruction aligns the transferred controls with the causal denoiser without changing its pretrained generative parameters. We combine this objective with sequence-level adversarial supervision:
\begin{equation}
 \mathcal L_{\mathrm{FA}}=\lambda(n_D)\mathcal L_{x_0}+0.01\,\mathcal L_{\mathrm{adv}},
 \qquad
 \mathcal L_{\mathrm{adv}}=\mathbb E\big[\operatorname{softplus}(-D_\xi(\widetilde z_\rho^{\mathrm{fake}},\rho;I^0,C_{\mathrm{geo}}))\big],
\label{eq:fast-loss}
\end{equation}
where $n_D$ counts discriminator updates and $\lambda(n_D)$ gradually increases the reconstruction weight.

\paragraph{Conditional sequence supervision.}
The video critic $D_\xi$ evaluates the generated interaction jointly with the initial image and rendered geometry. A frozen video backbone supplies ordered spatiotemporal features to learned register tokens, attention heads, and a clip classifier. Thus, the critic supervises the complete interaction sequence rather than independent frames. Let $\widehat z_0$ concatenate the supplied initial latent and the predicted endpoints. Real and generated clips share the same noise $\eta\sim\mathcal N(0,I)$ and noise level $\rho$:
\begin{equation}
 \widetilde z_\rho^{\mathrm{real}}=(1-\rho)z_0+\rho\eta,
 \qquad
 \widetilde z_\rho^{\mathrm{fake}}=(1-\rho)\widehat z_0+\rho\eta.
\label{eq:critic-noise}
\end{equation}
We train the discriminator with a logistic objective and use the non-saturating generator loss in Eq.~\ref{eq:fast-loss}. During generator updates, critic parameters remain fixed while gradients propagate to the generated latents through its frozen backbone.

\paragraph{On-policy autoregressive distillation.}
Condition alignment uses clean historical latents, whereas inference conditions on generated history. Following OPSD-V~\citep{liu2026opsd}, the student performs the deployed two-step rollout and commits its generated blocks to its own causal cache. An exponential-moving-average teacher evaluates the same student-visited noisy latent and denoising time with a cleaner autoregressive cache: the most recent generated block is retained, while older history is replaced by clean target latents. For student and teacher velocities $v^S_{s,k}$ and $v^T_{s,k}$ at block $s$ and denoising step $k$, we optimize
\begin{equation}
 \mathcal L_{\mathrm{OPD}}=\frac{1}{SK}\sum_{s=1}^{S}\sum_{k=1}^{K}
 \big\|v^S_{s,k}-\operatorname{sg}(v^T_{s,k})\big\|_2^2,
 \qquad
 \mathcal L_{\mathrm{post}}=\mathcal L_{\mathrm{OPD}}+0.02\,\mathcal L_{\mathrm{adv}},
\label{eq:opd-loss}
\end{equation}
with $K=2$. Rank-128 LoRA~\citep{hu2022lora} adapts the backbone. Local velocity targets correct student-visited denoising states, while the critic evaluates the resulting generated-history sequence. Both stages retain the same denoising schedule and cache update rule.

\paragraph{Bounded-memory gradient replay.}
For condition alignment, retaining all causal-block graphs together with the sequence critic is memory intensive. We first form the endpoint clip without retaining the generator graph and obtain the adversarial derivatives $g_s=\partial\mathcal L_{\mathrm{adv}}/\partial\widehat z_{0,s}$ from a detached latent leaf. For trainable generator parameters $\psi$, we then replay each block with the same noise and detached clean history, accumulating
\begin{equation}
 \nabla_\psi\mathcal L_{\mathrm{FA}}
 =\lambda(n_D)\nabla_\psi\mathcal L_{x_0}
 +0.01\sum_{s=1}^{S}
 \left(\frac{\partial\widehat z_{0,s}}{\partial\psi}\right)^{\!\top}g_s.
\label{eq:gradient-replay}
\end{equation}
The appearance encoder similarly accumulates derivatives through a cached feature leaf and is recomputed to update its parameters. This preserves gradients to the transferred conditioning branches without retaining their graphs across all causal queries. Appendix~\ref{sec:fast-objective-details} gives the critic architecture, optimization settings, and replay implementation.

\paragraph{Autoregressive inference.}
The renderer generates eight four-frame blocks with two denoising steps per block, conditioning on the initial observation and its generated visual history. Only a completed clean block is committed to the backbone and conditioning caches; noisy denoising queries do not append cache entries. For a fixed supplied action and camera trajectory, 3D forecasting and geometric projection are performed once for the rollout, and their controls are reused during subsequent latent-block generation. Future RGB observations are not used for appearance conditioning or generation history. The causal hand and appearance encoders access only the current and preceding control chunks.

\section{Experiments}
\begin{figure}[!t]
\centering
\includegraphics[width=0.9\linewidth]{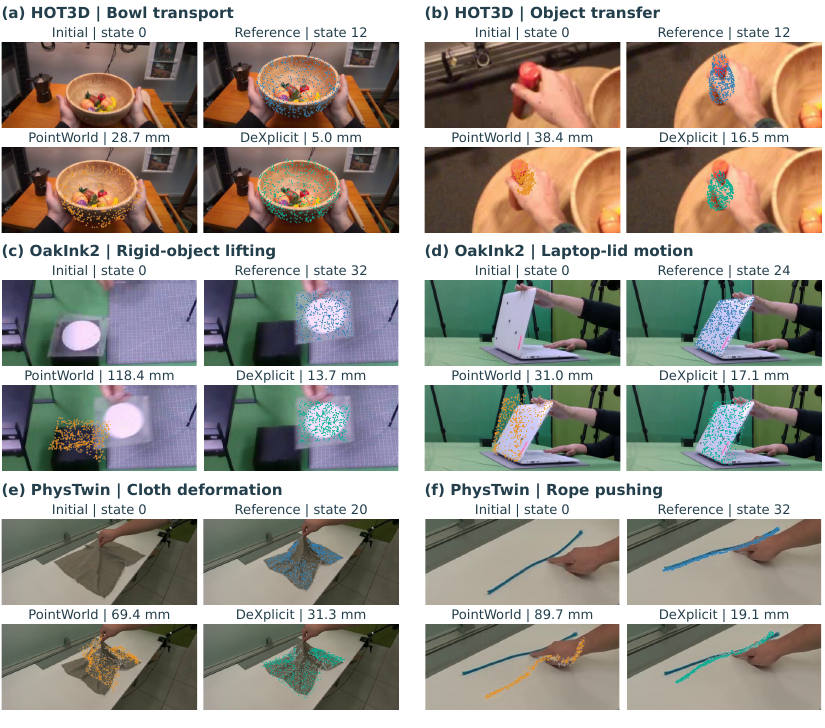}
\caption{\textbf{Qualitative 3D dynamics.} Six examples span rigid transport, articulation, and deformation. PointWorld and \method{} forecasts are overlaid on the same reference image; EPE is measured on reference-supported object points in millimeters.}
\label{fig:clean-native-3d}
\end{figure}

\Cref{fig:clean-native-3d} shows the corresponding interaction forecasts: shared transport preserves coherent motion, while local deformation captures articulation and cloth shape change.

\label{sec:experiments-dexsim}
\subsection{Experimental Setup}
\paragraph{Datasets and baselines.}
We evaluate one dynamics model across HOT3D, OakInk2, and PhysTwin~\citep{banerjee2025hot3d,yang2024oakink2,jiang2025phystwin}, spanning rigid, articulated, and deformable interactions. We retrain PointWorld on the same forecasting task using the same initial metric scene points and 3D hand-joint trajectories through its native point-action interface. On HOT3D, video systems use their native controls. \method{} predicts future object geometry, as with PointWorld (Full protocols in \Cref{sec:hot48}). All temporal windows derived from the same source unit remain in the same partition, preventing shared source interactions between our task-specific training and evaluation data. Within each comparison panel, all methods are evaluated on the same fixed set of source windows and reference targets.

\paragraph{Evaluation metrics.}
For 3D forecasting, we report moved-point and all-point errors at early and long horizons. Video metrics include PSNR, SSIM~\citep{wang2004ssim}, LPIPS~\citep{zhang2018lpips}, hand PCK20, object overlap and position error, and CLIP foreground similarity~\citep{radford2021clip}. The initial frame is excluded; cohorts, missing-prediction handling, support, and metric definitions are detailed in \Cref{sec:hot48}.

\subsection{Quantitative and Qualitative Results}
\paragraph{3D dynamics prediction.}
As shown in \Cref{tab:clean-geometry}, \method{} reduces moved-point error on all three datasets by 42--48\% at the early horizon and 27--40\% at the long horizon, demonstrating consistently more accurate prediction of interaction-induced geometric changes. All-point error also improves on HOT3D and PhysTwin, but is higher on OakInk2 despite the lower moved-point error.

\begin{table}[t]
\centering
\caption{\textbf{Metric 3D forecasting.} Both methods receive the same initial metric scene points and 3D hand-joint trajectories through their respective conditioning interfaces. Moved-point and all-point L2 (mm) with identical scene and action inputs. Errors are weighted by the corresponding valid point-time counts.}
\label{tab:clean-geometry}
\setlength{\tabcolsep}{5pt}\renewcommand{\arraystretch}{1.08}
\begin{center}
\resizebox{0.9\linewidth}{!}{
\begin{tabular*}{\linewidth}{@{}l@{\extracolsep{\fill}}rr|rr|rr@{}}
\toprule
&\multicolumn{2}{c|}{HOT3D}&\multicolumn{2}{c|}{OakInk2}&\multicolumn{2}{c}{PhysTwin}\\
\cmidrule(lr){2-3}\cmidrule(lr){4-5}\cmidrule(lr){6-7}
Model & Moved & All & Moved & All & Moved & All\\
\midrule
\multicolumn{7}{@{}l}{\textbf{(a) Long horizon: states 1--32}}\\
PointWorld & 90.47 & 18.17 & 83.21 & \textbf{11.71} & 60.41 & 35.52\\
\method{} & \textbf{54.34} & \textbf{14.01} & \textbf{59.95} & 16.88 & \textbf{44.04} & \textbf{25.37}\\
\midrule
\multicolumn{7}{@{}l}{\textbf{(b) Early horizon: states 1--14}}\\
PointWorld & 70.00 & 12.16 & 35.80 & \textbf{6.70} & 41.71 & 23.80\\
\method{} & \textbf{39.21} & \textbf{8.65} & \textbf{20.61} & 8.01 & \textbf{21.81} & \textbf{13.60}\\
\bottomrule
\end{tabular*}
}
\end{center}
\par\vspace{4pt}
\end{table}

\paragraph{Conditional video generation.}
\label{sec:hoi}
\method{} achieves the strongest perceptual quality and object overlap in the shared video panel. Baselines with higher PCK20 but much lower object overlap show that hand agreement alone does not capture interaction fidelity.

\begin{table}[t]\centering
\vspace{-0.5cm}
\caption{\textbf{Video generation.} Perceptual quality, hand alignment, object fidelity, and latency across output grids and horizons. PA is aligned hand-shape error; PCK20 uses a 20-pixel threshold. IoU$_+$ and OPE$_+$ penalize missing predictions; CLIP is reported with foreground coverage (Cov.). Full metric, support, and timing protocols: \Cref{sec:expanded-task-metrics,sec:causal-speed-only}}
\label{tab:hot48-main}
\begingroup\setlength{\tabcolsep}{1.4pt}
\begin{center}
\resizebox{0.9\linewidth}{!}{
\begin{tabular*}{\linewidth}{@{}l@{\extracolsep{\fill}}rrrrrrrrrr@{}}
\toprule
 & \multicolumn{3}{c}{Perceptual} & \multicolumn{2}{c}{Hand} & \multicolumn{4}{c}{Object} & Inference\\
\cmidrule(lr){2-4}\cmidrule(lr){5-6}\cmidrule(lr){7-10}\cmidrule(lr){11-11}
Method & PSNR & SSIM & LPIPS & PCK20 & PA & IoU$_+$ & OPE$_+$ & CLIP & Cov. & Speed\\
 & $\uparrow$ & $\uparrow$ & $\downarrow$ & (\%)$\uparrow$ & (mm)$\downarrow$ & $\uparrow$ & $\downarrow$ & $\uparrow$ & (\%)$\uparrow$ & (s)$\downarrow$\\
\midrule
\multicolumn{11}{l}{\textit{(a) Video comparison: future frames 1--32, $480\times832$}}\\
DeXplicit & \textbf{19.13} & \textbf{0.652} & \textbf{0.274} & 40.3 & 13.88 & \textbf{0.549} & \textbf{0.352} & \textbf{0.880} & 78.2 & 277.44\\
Causal-DeXplicit & 17.41 & 0.583 & 0.348 & 44.7 & 12.73 & 0.490 & 0.365 & 0.863 & \textbf{84.2} & 4.90\\
PointWorld (adapted) & 18.01 & 0.616 & 0.315 & 27.3 & 15.35 & 0.373 & 0.488 & 0.847 & 69.6 & 291.18\\
Hand2World & 14.32 & 0.505 & 0.497 & 54.9 & 12.04 & 0.082 & 0.864 & 0.835 & 22.1 & 15.76\\
Hand2World-AR & 13.47 & 0.487 & 0.604 & 57.0 & \textbf{11.97} & 0.080 & 0.861 & 0.836 & 22.2 & 4.62\\
InterDyn & 14.28 & 0.482 & 0.539 & \textbf{58.9} & 12.23 & 0.131 & 0.552 & 0.809 & 72.7 & 96.02\\
DWM (adapted) & 12.52 & 0.450 & 0.637 & 7.7 & 16.68 & 0.046 & 0.737 & 0.794 & 51.7 & 172.48\\
RynnWorld-SFT & 14.72 & 0.497 & 0.528 & 19.1 & 15.43 & 0.137 & 0.577 & 0.817 & 69.0 & 16.93\\
RynnWorld-Causal & 14.70 & 0.506 & 0.566 & 25.8 & 14.93 & 0.097 & 0.655 & 0.807 & 61.8 & 4.33\\
FlashMotion & 14.62 & 0.476 & 0.528 & 17.9 & 14.44 & 0.178 & 0.443 & 0.832 & 77.2 & \textbf{3.44}\\
\midrule
\multicolumn{11}{l}{\textit{(b) Native Mask2IV comparison: future frames 1--15, $320\times512$}}\\
DeXplicit & \textbf{20.88} & \textbf{0.700} & \textbf{0.189} & \textbf{43.9} & \textbf{12.66} & \textbf{0.580} & \textbf{0.329} & \textbf{0.884} & \textbf{77.7} & 277.44\\
Mask2IV & 14.39 & 0.463 & 0.479 & 16.9 & 14.53 & 0.166 & 0.545 & 0.810 & 67.4 & \textbf{26.05}\\
\midrule
\multicolumn{11}{l}{\textit{(c) Native future-image comparison: endpoint 32, $256\times256$}}\\
DeXplicit & \textbf{16.66} & \textbf{0.533} & \textbf{0.248} & 48.0 & 14.19 & \textbf{0.525} & \textbf{0.354} & \textbf{0.853} & \textbf{80.0} & 277.44\\
CosHand & 12.49 & 0.328 & 0.500 & \textbf{63.5} & \textbf{13.25} & 0.050 & 0.751 & 0.737 & 57.5 & \textbf{2.30}\\
\bottomrule\end{tabular*}
}
\end{center}
\par\endgroup\end{table}

\begin{figure}[!t]
\centering
\includegraphics[width=\linewidth]{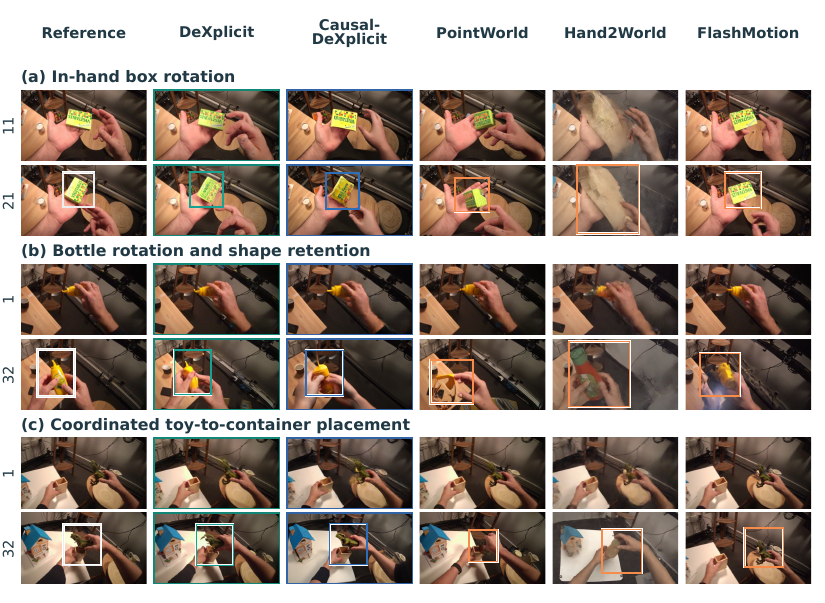}
\caption{\textbf{Qualitative video generation.} Bounding boxes highlight the manipulated object at the later frame. \method{} preserves object identity, shape, and appearance. Evaluation scope is in \Cref{sec:hot48}.}
\label{fig:main-video-comparison}
\end{figure}

\Cref{fig:main-video-comparison} shows object orientation and identity during rotation and placement; Causal-\method{} retains these properties under generated-history conditioning.

\paragraph{Generation efficiency.}
\label{sec:efficient-generation}
Causal-\method{} generates four frames in 0.54 seconds and all 32 in 4.90 seconds, with substantially higher object overlap than similarly fast autoregressive baselines (\Cref{tab:hot48-main}) operating at comparable latency, demonstrating a favorable trade-off between generation efficiency and object-level consistency. These results further validate our explicit-3D-first design: by resolving object dynamics in lightweight 3D prediction before visual synthesis, the renderer can produce geometrically consistent future observations without relying on expensive iterative video generation.

\subsection{Ablation Studies}
\label{sec:main-ablations}

\begin{table}[!t]
\centering
\begin{minipage}[t]{0.485\linewidth}
\centering
\caption{\textbf{Dynamics ablation.} L2 error (mm), states 1--34.}
\label{tab:dynamics-progression}
\setlength{\tabcolsep}{2pt}
\begin{tabularx}{\linewidth}{@{}>{\raggedright\arraybackslash}Xrr@{}}
\toprule
Configuration & Moved$\downarrow$ & All$\downarrow$\\
\midrule
Direct points & 63.346 & 13.579\\
+ action differences & 59.836 & 13.098\\
+ joint-aware encoding & 56.176 & 11.758\\
+ structured transport & 51.082 & 9.502\\
+ metric-aligned supervision & 50.182 & 9.358\\
+ canonical local graph & \textbf{49.254} & \textbf{9.226}\\
\bottomrule
\end{tabularx}
\end{minipage}%
\hfill
\begin{minipage}[t]{0.485\linewidth}
\centering
\caption{\textbf{Video conditioning ablation.}}
\label{tab:video-progression-main}
\setlength{\tabcolsep}{2pt}
\begin{tabularx}{\linewidth}{@{}>{\raggedright\arraybackslash}Xrrr@{}}
\toprule
Configuration & PSNR$\uparrow$ & SSIM$\uparrow$ & LPIPS$\downarrow$\\
\midrule
\multicolumn{4}{@{}l}{\emph{5B geometry study}}\\
Hand conditioning & 14.08 & .465 & .543\\
+ scene tracks & \textbf{16.13} & \textbf{.535} & \textbf{.409}\\
\midrule
\multicolumn{4}{@{}l}{\emph{1.3B conditioning stages}}\\
Geometry & 17.28 & .555 & .344\\
+ hand conditioning & 18.53 & .595 & .294\\
+ appearance transport & \textbf{18.81} & \textbf{.597} & \textbf{.289}\\
\bottomrule
\end{tabularx}
\end{minipage}
\end{table}


\Cref{tab:dynamics-progression,tab:video-progression-main} show that structured transport gives the largest dynamics gain, while scene tracks, hand conditioning, and appearance transport progressively improve video metrics.

\paragraph{Causal post-training.}
\Cref{tab:causal-posttraining} compares on-policy continuations from the same condition-aligned starting model, with identical training data, update budget, and low-rank backbone adaptation. The conditional video critic is the controlled difference. Relative to distillation alone, adding the critic reduces LPIPS from 0.377 to 0.348 and improves object IoU$_+$ from 0.459 to 0.490, together with higher PCK20 and object-pair coverage and lower OPE$_+$. These results support combining local denoising targets with sequence-level interaction supervision while retaining the same two-step inference procedure. Complete metrics and qualitative comparisons are provided in Appendix~\ref{sec:opsd-results}.

\begin{table}[htbp]
\centering
\caption{\textbf{Causal post-training.} Matched on-policy continuations from Fast Adapter, evaluated on the held-out causal cohort. PCK20 and coverage are percentages. All variants retain the same two-step sampler.}
\label{tab:causal-posttraining}
\begingroup\setlength{\tabcolsep}{3.2pt}
\begin{center}
\begin{tabular*}{\linewidth}{@{}l@{\extracolsep{\fill}}rrrrr@{}}
\toprule
Method & LPIPS$\downarrow$ & PCK20$\uparrow$ & IoU$_+\uparrow$ & OPE$_+\downarrow$ & Cov.$\uparrow$\\
\midrule
Fast Adapter & 0.370 & 41.88 & 0.460 & 0.392 & 80.7\\
On-policy distillation & 0.377 & 40.72 & 0.459 & 0.411 & 79.1\\
On-policy distillation + critic & \textbf{0.348} & \textbf{44.72} & \textbf{0.490} & \textbf{0.365} & \textbf{84.2}\\
\bottomrule
\end{tabular*}
\end{center}
\par\endgroup
\end{table}

\paragraph{Object geometry and appearance conditioning.}
We additionally evaluate the contribution of explicit object controls to the causal renderer (\Cref{tab:causal-object-controls}). With the same frozen renderer weights and inference seed, we remove projected object geometry and its associated appearance transport while retaining the initial observation, supplied hand conditioning, cameras, and native edit mask. Removing these object controls reduces PSNR from 16.744 to 14.582 and object IoU$_+$ from 0.444 to 0.248, while increasing LPIPS from 0.390 to 0.506. This is an inference-time intervention on a retained causal checkpoint, separate from the post-trained model in \Cref{tab:causal-posttraining}; it measures the joint contribution of object geometry and transported appearance rather than either component in isolation. Both variants use predicted geometry where object controls are enabled. The diagnostic comprises 48 windows from 11 source clips, of which 45 windows overlap the retained 3D predictor's training inventory, and therefore evaluates control dependence rather than held-out dynamics generalization.

\begin{table}[htbp]
\centering
\caption{\textbf{Object conditioning in the causal renderer.} Paired inference-time interventions on a retained causal checkpoint over 48 windows. Renderer weights, seed, and two-step sampling are fixed. The second row removes both projected object geometry and its transported appearance; hand conditioning is retained.}
\label{tab:causal-object-controls}
\begin{tabular*}{\linewidth}{@{}l@{\extracolsep{\fill}}rrr@{}}
\toprule
Configuration & PSNR $\uparrow$ & LPIPS $\downarrow$ & IoU$_+$ $\uparrow$\\
\midrule
Predicted object geometry + appearance transport & \textbf{16.744} & \textbf{0.390} & \textbf{0.444}\\
w/o object geometry and appearance transport & 14.582 & 0.506 & 0.248\\
\bottomrule
\end{tabular*}
\end{table}

To examine appearance transport separately, \Cref{tab:appearance-parent} compares the hand-conditioned 1.3B model with its appearance-enhanced continuation on the same 48 held-out windows under predicted geometry. Appearance transport improves PSNR from 18.3607 to 18.5658, SSIM from 0.6188 to 0.6329, and LPIPS from 0.3003 to 0.2835. This comparison uses a separate 1.3B bidirectional renderer and is distinct from the causal control intervention above.

\begin{table}[htbp]
\centering
\caption{\textbf{Effect of appearance transport.} The hand-conditioned model and its appearance-enhanced continuation are evaluated on 48 held-out windows with predicted geometry and the same 1.3B backbone. }
\label{tab:appearance-parent}
\begin{center}
\begin{tabular*}{\linewidth}{@{}l@{\extracolsep{\fill}}rrr@{}}
\toprule
Configuration & PSNR $\uparrow$ & SSIM $\uparrow$ & LPIPS $\downarrow$\\
\midrule
Geometry + hand conditioning & 18.3607 & 0.6188 & 0.3003\\
+ appearance transport & \textbf{18.5658} & \textbf{0.6329} & \textbf{0.2835}\\
\bottomrule
\end{tabular*}
\end{center}
\end{table}

\section{Conclusion}
We presented \method, a dexterous manipulation world model built around a simple principle: predict how the world changes explicitly in 3D before synthesizing how those changes look. By separating structured metric dynamics from visual synthesis, our approach provides geometrically grounded future states while reducing the burden on the video generator to implicitly learn action-conditioned world dynamics. The predicted 3D states further provide a strong scaffold for efficient causal rendering, enabling coherent future observations with substantially reduced generation cost. 

\label{main:end}

\subsection*{Reproducibility Statement}
We provide detailed evaluation sets, protocols, prediction horizons, baseline interfaces, metric definitions, and missing-prediction handling in Appendix~\ref{sec:hot48}. Dynamics architecture and optimization details are provided in Appendix~\ref{sec:dynamics-implementation}, video-generation architecture and inference settings in Appendix~\ref{sec:wan13b-details}, and causal adaptation objectives and training procedures in Sec.~\ref{sec:fast-adaptation} and Appendix~\ref{sec:fast-objective-details}. Additional controlled ablations are reported in Appendix~\ref{sec:component-studies}. Together, these sections specify the data, model architecture, training, evaluation, and inference procedures used to obtain the reported results.

\subsection*{ETHICS STATEMENT}
The experiments concern visual prediction for manipulation. Plausible generated futures can mis-
represent physical outcomes. The reported evaluations do not establish suitability for safety critical
robotic control.

\subsection*{AI Assistance}
AI tools assisted with software implementation, experiment and evaluation. All AI-assisted code were reviewed by the authors, who take responsibility for the final content and reported results.
\begingroup
\urlstyle{same}
\bibliographystyle{plainnat}
\bibliography{references,additional_references}
\endgroup
\clearpage
\appendix
\section{Evaluation Details}
\label{sec:hot48}\label{sec:expanded-task-metrics}\label{sec:baseline-details}

For HOT3D, split membership is assigned by source clip rather than independently by window. The video split construction preserves the canonical evaluation clip list while allocating development subsets from the remaining clips.

For OakInk2, split assignment is determined from the source sequence identifier before temporal windows are exported, and manifest construction rejects sequences shared across partitions. For PhysTwin, split assignment precedes both temporal window extraction and camera-view expansion, so different views of the same interaction cannot enter different partitions.

Video comparisons use a shared evaluation population with explicit window and source identifiers. Input and reference-file hashes are recorded, and the evaluation pipeline checks generated outputs against the declared population. Missing generated files trigger an error rather than silently reducing a method’s evaluation set. 

\paragraph{Evaluation sets.}
The dynamics datasets contain 2,720 training and 302 evaluation windows for HOT3D, 9,010 training and 1,127 evaluation windows for OakInk2, and 91 training and 76 evaluation windows for PhysTwin. \method{} and retrained PointWorld receive identical initial geometry, hand actions, and cameras in the 3D forecasting comparison.

\paragraph{Conditional video inputs and qualitative scope.}
The video comparison uses method-specific controls. Methods retain their native conditioning interfaces, model capacities, and inference procedures. The 14B \method{} results use predicted future object geometry and learned video adapters. PointWorld supplies predicted point trajectories and uses the same renderer. These video outputs evaluate the respective conditional systems. The causal model uses predicted geometry and hand conditioning and generates its own video history. The main RGB examples are selected illustrations. Additional appendix examples use the same 14B renderer. Geometry forecasting results remain separate from these conditional-rendering studies. 

\paragraph{Retrained PointWorld baseline.}
We retrain PointWorld~\citep{huang2026pointworld} on the same hand-object prediction setting. Its original action representation propagates robot-gripper surface points through a supplied joint trajectory using forward kinematics. We retain this point-action interface and substitute the supplied hand trajectories $H\in\mathbb{R}^{T\times N_H\times3}$. The baseline therefore receives the same metric action information through its original conditioning pathway. Initial scene coordinates and scene features, including color, normals, visibility, and precomputed DINO features and hand-point trajectory are supplied across the prediction horizon. Training follows the public PointWorld implementation with the same training and evaluation samples. \Cref{tab:pointworld_retrain} summarizes the configuration.

\begin{table}[t]
\centering

\caption{Retrained PointWorld baseline. The architecture and point-action
conditioning follow PointWorld, with robot trajectory points replaced by
3D hand trajectories.}
\label{tab:pointworld_retrain}
\begin{center}
\begin{tabular}{ll}
\toprule
Setting & PointWorld baseline \\
\midrule
Backbone & PointTransformerV3 \\
Scene input & Same initial metric points as \method{} \\
Action input & $H\in\mathbb{R}^{T\times N_H\times3}$ hand trajectories \\
Action fusion & Native PointWorld point-action conditioning \\
Structured joint encoder & None \\
Object transport and local graph & None \\
Training data & Same training split as \method{} \\
Prediction horizon & Same as \method{} \\
Selection metric & moved-point L2 on validation set \\
\bottomrule
\end{tabular}
\end{center}
\end{table}

\paragraph{Causal evaluation.}
The causal model generates eight four-frame blocks from the initial observation, using two denoising steps and guidance scale 1. Each completed latent block enters the cache for subsequent prediction. Future reference images are excluded from the generation history. All 32 generated frames are scored at $480\times832$. Metrics requiring successful estimates are distinguished from those that explicitly penalize missing predictions.

\subsection{Causal Post-Training Evaluation}
\label{sec:opsd-results}
The matched post-training comparison is reported in Sec.~\ref{sec:main-ablations}. Both continuations share the condition-aligned starting model, low-rank adaptation, training data, and update budget; the video critic is the controlled difference.

\begin{figure}[!t]
\centering
\includegraphics[width=\linewidth]{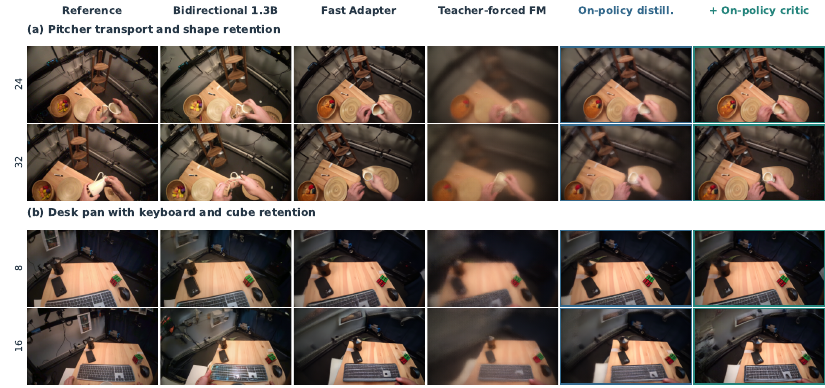}
\caption{\textbf{Ablation of causal post-training.} On-policy distillation with the video critic preserves object shape and placement in two held-out examples. The bidirectional model provides a non-causal reference.}
\label{fig:causal-posttraining-qualitative}
\end{figure}

\Cref{fig:causal-posttraining-qualitative} illustrates the effect of the video critic on object shape and placement. The matched quantitative comparison appears in \Cref{tab:causal-posttraining}.

\Cref{tab:causal-posttraining-full} reports the remaining image and interaction metrics for the same evaluated variants.
\begin{table}[t]
\centering
\caption{\textbf{Complete causal post-training metrics.} Evaluated variants under the same two-step inference protocol. PA is aligned hand error in millimeters. PCK20 and coverage are percentages.}
\label{tab:causal-posttraining-full}
\begingroup\setlength{\tabcolsep}{1.6pt}
\begin{center}
\begin{tabular*}{\linewidth}{@{}l@{\extracolsep{\fill}}rrrrrrrrr@{}}
\toprule
Method & PSNR$\uparrow$ & SSIM$\uparrow$ & LPIPS$\downarrow$ & PCK20$\uparrow$ & PA$\downarrow$ & IoU$_+\uparrow$ & OPE$_+\downarrow$ & CLIP$\uparrow$ & Cov.$\uparrow$\\
\midrule
Fast Adapter & 17.19 & .582 & .370 & 41.88 & 13.41 & .460 & .392 & .854 & 80.7\\
On-policy distillation & 17.30 & \textbf{.595} & .377 & 40.72 & 13.17 & .459 & .411 & .856 & 79.1\\
+ video critic & \textbf{17.41} & .583 & \textbf{.348} & \textbf{44.72} & \textbf{12.73} & \textbf{.490} & \textbf{.365} & \textbf{.863} & \textbf{84.2}\\
\bottomrule
\end{tabular*}
\end{center}\endgroup
\end{table}

\paragraph{Prediction horizons.}
Metric dynamics use context state 0 and 34 future states. Early and full moved-point errors cover states 1--14 and 1--34. The main cross-dataset geometry table uses states 1--14 and 1--32. Endpoint error uses ever-moved points at state 34. Persistent point identities and reference validity determine correspondence. Distances aggregate by valid moved point-time counts, as in the main-paper definition. Video evaluation uses frames 1--32 at $480\times832$, excluding the supplied initial frame, and averages equally over windows.

\paragraph{Baseline inference.}
PointWorld supplies forecast geometry to a video diffusion renderer. Hand2World uses hands and cameras. RynnWorld uses depth-aware skeletons, InterDyn uses amodal hand masks, and FlashMotion uses hand and object boxes. DWM denotes the retrained skeleton-control adaptation with static-scene conditioning. Mask2IV uses an initial image and mask with generic text, while CosHand uses initial and query hand silhouettes. DexWM supplies latent features and a native hand probe. 

\paragraph{Video and hand metrics.}
RGB quality uses PSNR, Gaussian-window SSIM~\citep{wang2004ssim}, and AlexNet LPIPS~\citep{zhang2018lpips}. PCK at 10 and 20 pixels scores reference-valid, in-image joints, with missing predictions counted as incorrect. Annotation targets and independently extracted reference hands are evaluated separately. Wrist-relative MPJPE removes wrist translation. PA-MPJPE additionally aligns rotation and scale. Reconstruction errors are computed on successful fits.

\paragraph{Object metrics.}
IoU$_+$ measures overlap on reference-positive object masks, assigning zero to missing predictions. OPE$_+$ measures mask-centroid distance normalized by image diagonal, assigning one to missing predictions. Masked object CLIP~\citep{radford2021clip} uses paired foreground and is reported with pair coverage. OOE measures the image-plane angle between mask principal axes. 

\paragraph{Native output grids.}
Mask2IV is evaluated at its native $16\times320\times512$ output, scoring frames 1--15. CosHand predicts the endpoint image at $256\times256$, with the same centered crop applied to predictions, references, and projected annotations. It uses the released $S=50$ DDIM schedule, guidance 1.5, and $\eta=1$. Each native panel uses its own spatial grid and horizon.

\section{Additional Quantitative Results}
\begin{table}[htbp]
\centering
\caption{\textbf{DexWM native hand probe.} Percent correct using the native $224\times392$ grid and 12 temporal slots. PCK10 at the final slot is 2.43\%.}
\label{tab:dexwm-matched}
\begin{center}
\begin{tabular*}{\linewidth}{@{}l@{\extracolsep{\fill}}rrrr@{}}
\toprule
Target support & PCK2.5 & PCK5 & PCK10 & PCK20\\
\midrule
Native fixed slots (18,432) & 0.72 & 2.05 & 7.06 & 20.42\\
Valid annotated slots (10,836) & 1.22 & 3.48 & 11.94 & 33.71\\
\bottomrule
\end{tabular*}
\end{center}
\end{table}

\clearpage
\section{Implementation Details}
\label{sec:wan13b-details}\label{sec:video-conditioning}\label{sec:training-status}\label{sec:fast-details}
\subsection{Dynamics Architecture and Optimization}
\label{sec:dynamics-implementation}

This section specifies $F_\theta$ in Sec.~\ref{sec:structured}, including the architecture and optimization of the shared model used for cross-dataset evaluation.

\paragraph{Scene and action tensors.}
Each example contains one observed scene and a supplied 35-state action sequence, in which state 0 is context and states 1--34 are predicted jointly. The scene has at most 12,000 persistent points. Smaller clouds are padded and masked, while larger clouds are subsampled with consistent indices across time. Coordinates are expressed in meters without additional normalization. Only initial-state scene features enter the predictor. Precomputed pointwise DINO features and the initial geometric and appearance attributes are projected to width 256. Future object coordinates are supervision targets. The two hands retain 21 anatomical joints each. The joint encoder has four Transformer layers, width 256, four heads, feed-forward width 1,024, and dropout 0.1. The separate MANO-relative action encoder has two layers with the same width, head count, and dropout, using 15 pose coefficients and 10 shape coefficients per hand. Joint and action tokens are appended to the supplied hand-point tokens. Unavailable annotations are masked or zero-filled. Both absolute action states and backward differences are represented. For HOT3D, object membership follows the sampled object-mesh groups, and canonical coordinates are the corresponding object-local points. Legacy samples recover these coordinates using the inverse object pose at the context frame. MANO parameters are taken from the supplied hand annotations. These structured annotations are part of the declared input setting. PhysTwin provides sparse tracked hand-surface controls without reliable MANO joint identities. We retain these controls through the point-action interface.

\paragraph{Point Transformer and object-time network.}
The Point Transformer V3~\citep{wu2024ptv3} uses the base encoder--decoder configuration in \Cref{tab:dynamics-backbone}. Scene and action tokens have width 256. Spatial serialization uses a 0.015 m grid and the Z, transposed-Z, Hilbert, and transposed-Hilbert orders, with order shuffling enabled. Every attention patch has 256 points, and stage transitions have stride 2. The MLP expansion ratio is 4, attention uses biased query, key, and value projections, pre-normalization, zero attention and projection dropout, and maximum stochastic-depth rate 0.3. Relative positional attention is disabled. A two-layer, four-head temporal action Transformer at width 256 follows the point backbone. Masked mean and max object pooling and canonical-coordinate embeddings supply an eight-layer object-time Transformer with width 256, eight heads, feed-forward width 1,024, GELU, pre-normalization, and dropout 0.1. The pose head uses hidden width 128 and outputs three translation and six rotation coordinates. The latter are orthonormalized into a proper rotation~\citep{zhou2019rotation6d}. Dense displacement and log-variance heads also use hidden width 128 and predict all 34 future states.

\paragraph{Object-time fusion and rigid special case.}
Let $Z$ denote the pooled object tokens and $A$ the encoded hand action. The object-time Transformer produces
\begin{equation}
 U=\mathcal{T}_{\mathrm{obj}}(Z,A),\qquad
 U=\{U_o^t \mid o\in\mathcal O,\ t=1,\ldots,T\}.
\end{equation}
A pose head maps each $U_o^t$ to $(R_o^t,\tau_o^t)$. If local graph transforms are disabled, all points assigned to object $o_i$ share the rigid forecast
\begin{equation}
 \widehat x_i^t=R_{o_i}^t c_i+\tau_{o_i}^t.
\end{equation}
The main paper gives the more general global-plus-local deformation model.

\begin{table}[htbp]
\centering
\caption{\textbf{Point Transformer V3 configuration for dynamics.} Entries follow the stage order in the encoder and decoder configuration. Decoder stages execute from coarse to fine.}
\label{tab:dynamics-backbone}
\begin{center}
\begin{tabular}{@{}lll@{}}
\toprule
Setting & Encoder & Decoder\\
\midrule
Blocks per stage & $(4,4,4,8,8,12,4)$ & $(2,2,2,2,2,2)$\\
Channels & $(256,256,256,384,384,512,768)$ & $(256,256,256,384,384,512)$\\
Attention heads & $(4,4,4,8,8,16,24)$ & $(4,4,4,8,8,16)$\\
Patch size & 256 at every stage & 256 at every stage\\
\bottomrule
\end{tabular}
\end{center}
\end{table}

\paragraph{Canonical graph and spatial basis.}
We use $K=8$ nodes per object, selected by deterministic farthest-point sampling of valid canonical points, starting from the point farthest from their centroid. Objects with fewer points use the available nodes and mask the remainder. Each point is connected to its four nearest nodes and each node to four neighboring nodes. Skinning weights are proportional to $\exp(-\|c_i-g_k\|^2/(2r_o^2))$ and normalized over the selected nodes, where $r_o$ is 1.5 times the median nearest-node spacing, lower-bounded by 0.005 m. The spatial deformation rank is four: we use the four lowest-frequency nonconstant eigenvectors of the symmetric, unnormalized graph Laplacian. Edge affinities are Gaussian, with scale given by the median graph-edge length. Eigenvector signs are fixed by their largest-magnitude entries.

\paragraph{Temporal parameterization and deformation head.}
For the 34 future states, the six temporal controls define an open-uniform clamped cubic B-spline basis $B\in\mathbb R^{34\times6}$ on $u_t=(t-1)/33$, $t=1,\ldots,34$, with knot vector
\[
 (0,0,0,0,\tfrac13,\tfrac23,1,1,1,1).
\]
The network first predicts a six-dimensional twist for each spatial mode and future time. It projects these predictions into the temporal basis by $C=B^\dagger Z$ and reconstructs $\widetilde Z=BC$. The coefficient tensor has shape $6\times4\times6$ per object. Multiplication by the spatial basis yields node twists, whose node-mass-weighted mean is removed. The mode MLP has two hidden layers of width 256 with GELU and a zero-initialized six-channel output. Its inputs combine the object-time feature, pooled node feature, four node descriptors, a 17-dimensional action descriptor, and 18 node--hand relation features. Hand relations use a 0.04 m distance scale. A sigmoid deformability gate with hidden width 128 starts at 0.05. Gated componentwise $\tanh$ bounds restrict local translations to 0.03 m and rotation vectors to $12^\circ$ before the exponential map.

\paragraph{Pose composition and numerical precision.}
Dense predictions provide votes for a differentiable proper-Kabsch global pose, and graph deformation supplies the local residual. Assigned object points are rendered through the resulting global pose and canonical graph, while unassigned points retain the dense path. The gauge projection transfers the best-fitting rigid component of the warp into the global pose without changing predicted positions. The carrier branch scores five candidates (current pose, left hand, right hand, both hands, and initial pose), masks unavailable candidates, and uses width-128 score and strength heads with zero-initialized strength. Pose corrections use the 0.05 m and $15^\circ$ bounds specified below in Eq.~\ref{eq:pose-composition}. Training uses BF16 autocast with FP32 parameters and optimizer states. Rigid fitting, graph composition, and gauge operations run in FP32 with autocast disabled.

\paragraph{Training data.}
A single dynamics model is trained jointly across HOT3D, OakInk2, and PhysTwin using the training sets specified in \Cref{sec:hot48}. Training balances the larger datasets by uniformly resampling 1,000 HOT3D and 1,000 OakInk2 windows alongside all 91 PhysTwin windows. Sampling does not filter by motion magnitude.

\paragraph{Optimization.}
AdamW uses $(\beta_1,\beta_2)=(0.9,0.999)$, $\epsilon=10^{-8}$, weight decay 0.01, and gradient-norm clipping at 5. The graph, gate, and carrier heads use learning rate $6\times10^{-5}$. The pose head, bounded pose adapter, and final two object-time Transformer layers use $6\times10^{-6}$. Other parameters remain fixed during refinement. Active groups use linear warm-up followed by a constant phase and cosine decay to 10\% of their peak rates.

The objective retains point-flow uncertainty supervision and pose translation and rotation weights 20 and 2. The metric-aligned component weights are $(0.45,0.20,0.35)$, with smoothing $\epsilon=0.001$ m and a calibrated multiplier of 12.630234. Graph ARAP, temporal, magnitude, and gate weights are 0.25, 0.05, 0.01, and 0.10. Edge-strain matching has weight 0.5. Spline-coefficient, carrier temporal-variation, and carrier-strength penalties have weights $10^{-4}$, $10^{-3}$, and $10^{-4}$. The additional velocity, acceleration, horizon-position, object-displacement, and local-isometry losses are disabled in this stage.

\subsection{Geometric Objectives and Pose Refinement}
\paragraph{Pointwise uncertainty.}
For valid point-time pairs $\mathcal V$, the predictor learns log variance $s_i^t$ through
\begin{equation}
 \mathcal L_{\mathrm{point}}=\frac1{|\mathcal V|}\sum_{(i,t)\in\mathcal V}
 \left[e^{-s_i^t}\|\widehat x_i^t-x_i^t\|_2^2+s_i^t\right].
\end{equation}
The log variance is converted to bounded confidence for projection. The direct metric term combines full-horizon moved-point error, early moved-point error, and endpoint error on ever-moved points with weights 0.45, 0.20, and 0.35. Training uses the smooth distance $\sqrt{\|e\|_2^2+\epsilon^2}-\epsilon$, with $\epsilon=0.001$ m. Evaluation reports unsmoothed Euclidean distances.

\paragraph{Bounded object-frame corrections.}
A zero-initialized adapter predicts a local twist $(v,\omega)$ bounded componentwise by 0.05 m and $15^\circ$. Let $R_\Delta=\exp([\omega]_\times)$ and $t_\Delta=J(\omega)v$, where $J$ is the left Jacobian of $\mathrm{SO}(3)$. The corrected pose is
\begin{equation}
 R'=RR_\Delta,\qquad \tau'=\tau+Rt_\Delta.
\end{equation}
Both right-hand sides use the original pose. Hand-carried candidates come from rigid fits between initial and current hand joints. Candidate differences are expressed in the predicted object frame, bounded, and combined with learned normalized weights and a signed strength. Absent hands are masked. The current-pose candidate and zero strength allow the model to leave the object unchanged.

\paragraph{Gauge fixing.}
Global pose and local graph motion can encode the same rigid component. A weighted proper Kabsch transform~\citep{kabsch1976solution} $A$ fitted from canonical points to their warped positions is absorbed into the global pose $G$. Replacing $(G,Y)$ by $(GA,A^{-1}Y)$ preserves predicted world positions and leaves the graph responsible for residual deformation. This gives the pose refinement a consistent rigid reference.

\subsection{Video Generation}
\paragraph{Backbone and optimization.}
The bidirectional model uses the Wan2.2-VACE-14B backbone with rank-32 LoRA~\citep{hu2022lora} on VACE attention and feed-forward projections. The frozen backbone and VAE use BF16, while trainable adapters, encoders, and optimizer states use FP32. Training uses native full-range flow matching~\citep{lipman2023flowmatching,liu2023rectifiedflow}, AdamW at $10^{-4}$, and effective batch 21. Sampling uses 30 Euler steps, guidance 5, and shift 5 for $33\times480\times832$ video.

\paragraph{Hand conditioning.}
The hand branch combines 42 Gaussian joint heatmaps at $60\times104$ with a 642-dimensional per-frame vector. Each joint contributes current-camera and first-camera XYZ, wrist-relative XYZ, backward displacement, and three validity flags. A 12-value relative camera transform completes the vector. Four-frame packing yields nine temporal slots, repeating frame 0 for the initial group. The spatial encoder maps 168 channels through widths 128 and 1,536 with a $2\times2$ patch stride. The action MLP maps 2,568 inputs through width 256 to 1,536 and broadcasts spatially. Both output heads are zero-initialized.

\paragraph{Appearance transport.}
Persistent object points carry 16-channel VAE descriptors from the separately encoded initial image. Calibrated projection, closest-depth selection, and MANO occluders determine support, and a coverage channel marks valid features. Three temporal encoder stages of widths 256, 128, and 128 reduce 33 frames to nine slots, and a zero-initialized output supplies 96 VACE channels. The reference slot receives zero additional hand-conditioning or appearance-transport residual. The appearance-transport stream in \Cref{fig:wan13b-conditions} uses one PCA basis and color range per sequence, with unsupported cells shown in black.

\paragraph{Inference settings.}\label{sec:causal-speed-only}
\Cref{tab:streaming-latency} reports latency on A800 GPUs with resident model weights and native inference settings. After one full warm-up, we take the median of three synchronized, single-request measurements on a fixed evaluation window. Timing ends when decoded RGB is available on the host and excludes model loading, media storage, and metric computation. For causal models, the first completed block is decoded and transferred immediately. Full-output latency includes this early decode and the final complete-video decode. Other interfaces return the complete decoded output together. The 14B measurements include transfers between the high-noise and low-noise experts and VAE decoding.

Models retain their native input preparation. RynnWorld reuses input latents, while the other baselines use their original encoding procedures. Mask2IV generates 15 future frames and CosHand generates one future image. Their times are reported without extrapolation, whereas DeXplicit's native-grid comparisons use its complete 32-frame output. A separate batch-12 throughput measurement for Causal-DeXplicit, without early decoding, takes 47.36 seconds in total, or 3.95 seconds per video. This throughput measurement is distinct from single-request latency.

\begin{table}[t]\centering
\caption{\textbf{Autoregressive generation latency.} Time to the first decoded block and the complete 32-frame output.}
\label{tab:streaming-latency}
\begingroup\setlength{\tabcolsep}{8pt}
\begin{center}
\begin{tabular}{@{}lrr@{}}
\toprule
Method & First block (s) & Full output (s)\\
\midrule
Causal-DeXplicit & 0.54 & 4.90\\
Hand2World-AR & 0.31 & 4.62\\
RynnWorld-Causal & 0.62 & 4.33\\
\bottomrule
\end{tabular}
\end{center}
\par\smallskip\raggedright The first block contains 4 frames for Causal-DeXplicit and Hand2World-AR, and 12 for RynnWorld-Causal. Median single-request latency on A800 GPUs. Full-output time includes the early block decode. See \Cref{sec:causal-speed-only} for timing details.
\par\endgroup\end{table}

\section{Causal Adaptation Details}
\label{sec:fast-objective-details}
Section~\ref{sec:fast-adaptation} presents the causal adaptation objectives and inference procedure; \Cref{fig:multi-stage} summarizes the training stages. We provide the remaining architectural and optimization settings here.

\paragraph{Condition-alignment settings.}
In Eq.~\ref{eq:fast-loss}, $\lambda(n_D)=0.1+0.5\min(n_D/2000,1)$ increases the reconstruction weight with discriminator updates. Historical training blocks are detached. For critic comparisons in Eq.~\ref{eq:critic-noise}, we sample $u$ uniformly from $\{0,1/1000,\ldots,999/1000\}$ and set $\rho=\operatorname{clip}(5u/(1+4u),0.02,0.98)$.

\paragraph{Conditional video discriminator.}
The conditional video discriminator follows the Wan1.3B architecture. A frozen 30-block video backbone provides features from blocks 13, 21, and 29. Three learned register tokens and three attention heads aggregate these features together with text context, the initial image latent, and rendered trajectory latents. A classifier maps the concatenated register features to one clip logit. The heads use width 1,536, feed-forward width 8,960, and 12 attention heads. They operate on ordered video tokens, rather than scoring frames independently. Only registers, attention heads, and the classifier are trained.

Writing $d_r=D_\xi(\widetilde z_\rho^{\mathrm{real}},\rho;I^0,C_{\mathrm{geo}})$ and $d_f=D_\xi(\widetilde z_\rho^{\mathrm{fake}},\rho;I^0,C_{\mathrm{geo}})$, the discriminator objective is
\begin{equation}
 \mathcal L_D=0.01\,\mathbb E\big[\operatorname{softplus}(-d_r)+\operatorname{softplus}(d_f)\big].
 \label{eq:discriminator-loss}
\end{equation}
Both optimizers use AdamW with $\beta_1=0$, $\beta_2=0.999$, weight decay 0.01, and global gradient-norm clipping at 10.

\paragraph{Post-training settings.}
On-policy distillation uses rank-128 LoRA and the cleaner-history EMA teacher defined in Sec.~\ref{sec:fast-adaptation}. The inference schedule $1600/1601\rightarrow5/6\rightarrow0$, cache update rule, and two denoising calls per block are unchanged. The generated-history video critic has adversarial weight 0.02.

\paragraph{Gradient-replay implementation.}
In Eq.~\ref{eq:gradient-replay}, reconstruction is averaged over blocks, while the already reduced adversarial derivatives are accumulated without further normalization. This computes the gradient of the detached-history condition-alignment objective. The appearance encoder is evaluated once per minibatch to create a detached, gradient-bearing feature leaf. Block-wise replay accumulates the derivative with respect to that leaf; the small feature encoder is then recomputed once per sample to propagate the accumulated gradient into its parameters. Recomputed features are checked for exact agreement with the cached forward. Cached and replayed computations use the same precision. The BF16 autocast weight cache is disabled during replay to prevent reuse of detached parameter casts. Distributed workers complete their local causal replays before sample-weighted gradient synchronization, clipping, and the optimizer step.

\clearpage
\section{Additional Ablation Studies}
\label{sec:component-studies}
This section extends the dynamics and video ablations in \Cref{sec:main-ablations}. Comparisons retain the horizon, training budget, and evaluation set of each study.

\subsection{Action Encoding and Structured Object Motion}
In the shorter-horizon HOT3D encoder study, moved-point L2 decreases from 44.6 mm without the action encoder to 42.4 mm with a joint Transformer and to 36.4 mm with additional MANO-relative action tokens. These results support preserving articulated hand structure in the action representation. The subsequent 35-state study compares the action-conditioned point baseline with structured object-centric $\mathrm{SE}(3)$ transport, an object-time Transformer, and auxiliary losses. Endpoint error on ever-moved points decreases from 122.76 to 87.45 mm, a 28.8\% reduction for the structured configuration. The cumulative effect of the subsequent refinement components is evaluated in \Cref{tab:dynamics-progression}.

\subsection{Geometry, Hand Conditioning, and Appearance Transport}
\label{sec:component-video}
The geometry ablation uses Wan2.2-TI2V-5B with equal training budgets and 150 evaluation windows from 75 clips. The baseline conditions on hand pointmaps, and the comparison adds projected scene tracks. This study isolates the contribution of object geometry to hand-conditioned synthesis. The 1.3B study uses a separate VACE backbone with rendered geometry controls.

The hand and appearance rows in \Cref{tab:video-progression-main} represent successive conditioning stages. The matched appearance-transport comparison is reported in \Cref{tab:appearance-parent} in the main paper; both models use the same 48 held-out windows. The recordings do not overlap training. Appearance transport improves both pixel reconstruction and perceptual similarity, consistent with preserving correspondence to the initial observation. This comparison is separate from the main 14B and causal video evaluations.

\clearpage
\section{Additional Qualitative Results}
\label{sec:additional-dynamic-qualitative}
\begin{figure}[htbp]
\centering
\includegraphics[width=0.8\linewidth]{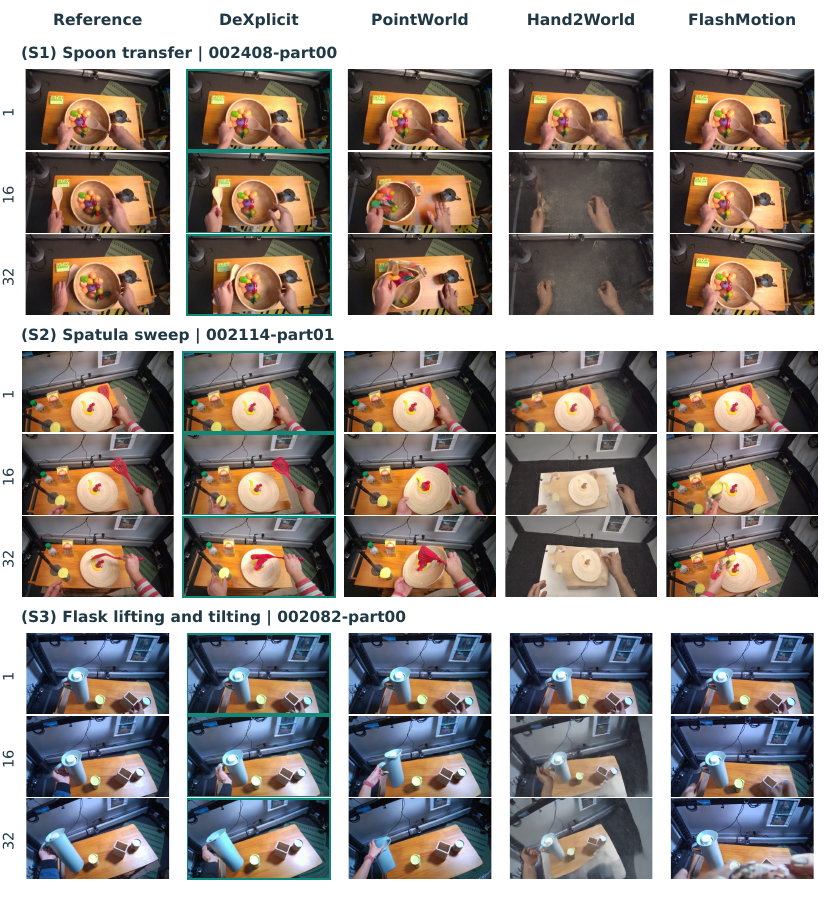}
\caption{\textbf{Dynamic interaction examples S1--S3.} Spoon transfer, spatula sweep, and flask lifting and tilting. Rows within each sample show future frames 1, 16, and 32. All methods use identical times and crops within each case. \method{} retains the manipulated object's appearance while following its changing placement and orientation. Baselines retain their native conditioning interfaces and generated outputs.}
\label{fig:additional-dynamic-videos-1}
\end{figure}

\begin{figure}[!t]
\centering
\includegraphics[width=0.8\linewidth]{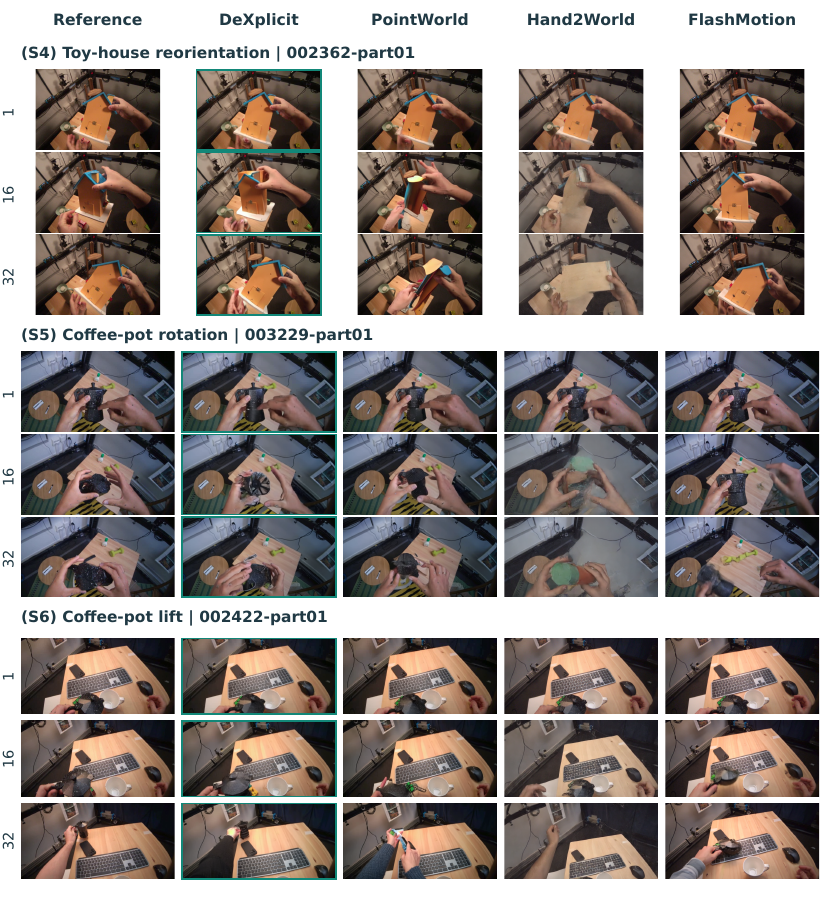}
\caption{\textbf{Dynamic interaction examples S4--S6.} Toy-house reorientation, coffee-pot rotation, and a coffee-pot lift in another recording. Future frames 1, 16, and 32 use the same crop for every method in each sample. The examples illustrate object identity and spatial response across rotation, changing visibility, and translation. }
\label{fig:additional-dynamic-videos-2}
\end{figure}

\Cref{fig:additional-dynamic-videos-1,fig:additional-dynamic-videos-2} show six further HOT3D windows from distinct source recordings, with no sample repeated from either main qualitative figure. Examples were selected using perceptual quality, annotated object motion, and the complete generated sequences. All six have more than 30\,mm maximum annotated object translation. The same 14B model is used throughout. These conditional-rendering illustrations use predicted future object geometry. They are selected qualitative examples, while quantitative tables retain the full evaluation cohorts. 

\clearpage

\FloatBarrier
\section{Limitations}
\label{sec:limitations}
\paragraph{Structured inputs.}
DeXplicit assumes an initial metric 3D scene and a future hand trajectory. Errors in these inputs may propagate into both physical-state prediction and video synthesis, and obtaining them reliably may require dedicated perception systems.

\paragraph{Unobserved appearance.}
Appearance transport can only propagate features visible in the initial observation. Newly revealed or heavily occluded regions must therefore be synthesized by the renderer. Although the confidence-gated visual readout explicitly handles unsupported regions, large disocclusions can still reduce visual fidelity.
\end{document}